\documentclass{BIGAI}
\usepackage{times}

\usepackage[colorlinks=true, citecolor=blue, linkcolor=black]{hyperref}
\usepackage[utf8]{inputenc} 
\usepackage[T1]{fontenc}    
\usepackage{booktabs}       
\usepackage{amsfonts}       
\usepackage{nicefrac}       
\usepackage{microtype}      
\usepackage{xcolor}         
\usepackage{url}            
\usepackage{bm}
\usepackage{amssymb}
\usepackage{amsmath}
\usepackage{graphicx}
\usepackage{epstopdf}
\usepackage{indentfirst}
\usepackage{wrapfig}
\usepackage{subfigure}
\usepackage{array}
\usepackage{color}
\usepackage{booktabs}
\usepackage{adjustbox}
\usepackage{makecell}
\usepackage{caption}
\usepackage{algorithm}
\usepackage{algorithmic}
\usepackage{mathtools}
\usepackage{amsthm}
\usepackage{enumitem}
\usepackage{mathptmx} 
\usepackage{fancyhdr} 
\usepackage{acronym}
\usepackage[capitalise,noabbrev,nameinlink]{cleveref}
\usepackage[most]{tcolorbox}
\usepackage{color}
\usepackage{longtable}
\usepackage{pgfplots}
\pgfplotsset{compat=1.3}
\usetikzlibrary{positioning, fit, calc}  
\tikzset{block/.style={draw, thick, text width=2cm ,minimum height=1.3cm, align=center},   
	line/.style={-latex}     
} 

\tikzset{blocktext/.style={draw, thick, text width=5.2cm ,minimum height=1.3cm, align=center},   
	line/.style={-latex}     
}

\tikzset{font=\footnotesize}

\definecolor{gblue}{HTML}{4285F4}
\definecolor{gred}{HTML}{DB4437}
\definecolor{ggreen}{HTML}{0F9D58}
\definecolor{gyellow}{HTML}{F4B400}

\acrodef{biv}[BiV]{Brain in a Vat}
\acrodef{llm}[LLM]{Large Language Model}
\acrodef{agi}[AGI]{artificial general intelligence}

\newcommand{\ie}{\textit{i}.\textit{e}.}
\newcommand{\eg}{\textit{e}.\textit{g}.}

\tcbset{
  aibox/.style={
    width=474.18663pt,
    top=10pt,
    colback=white,
    colframe=black,
    colbacktitle=black,
    enhanced,
    center,
    attach boxed title to top left={yshift=-0.1in,xshift=0.15in},
    boxed title style={boxrule=0pt,colframe=white,},
  }
}
\newtcolorbox{AIbox}[2][]{aibox,title=#2,#1}
\global

\crefname{algorithm}{Alg.}{Algs.}
\Crefname{algocf}{Algorithm}{Algorithms}
\crefname{section}{Sec.}{Secs.}
\Crefname{section}{Section}{Sections}
\crefname{table}{Tab.}{Tabs.}
\Crefname{table}{Table}{Tables}
\crefname{figure}{Fig.}{Figs.}
\Crefname{figure}{Figure}{Figures}
\crefname{equation}{Eq.}{Eqs.}
\Crefname{equation}{Equation}{Equations}
\crefname{appendix}{Appx.}{Appxs.}
\Crefname{appendix}{Appendix}{Appendices}

\title{Brain in a Vat: On Missing Pieces Towards Artificial General Intelligence in Large Language Models}

\author
{Yuxi Ma$^1$\footnote{Correspondence: mayuxi@bigai.ai}, Chi Zhang$^1$, Song-Chun Zhu$^{1,2}$\\
\vspace{1em} 
\normalfont{$^1$\small Beijing Insitute for General Artificial Intelligence (BIGAI)}\\
\normalfont{$^2$\small Peking University}
}

\begin{document}

\maketitle
\thispagestyle{firstpagestyle} 

\begin{abstract}
In this perspective paper, we first comprehensively review existing evaluations of \acfp{llm} using both standardized tests and ability-oriented benchmarks. We pinpoint several problems with current evaluation methods that tend to overstate the capabilities of \acp{llm}. We then articulate what artificial general intelligence should encompass beyond the capabilities of \acp{llm}. We propose four characteristics of generally intelligent agents: 1) they can perform unlimited tasks; 2) they can generate new tasks within a context; 3) they operate based on a value system that underpins task generation; and 4) they have a world model reflecting reality, which shapes their interaction with the world. Building on this viewpoint, we highlight the missing pieces in artificial general intelligence, that is, the unity of knowing and acting. We argue that active engagement with objects in the real world delivers more robust signals for forming conceptual representations. Additionally, knowledge acquisition isn't solely reliant on passive input but requires repeated trials and errors. We conclude by outlining promising future research directions in the field of artificial general intelligence.
\end{abstract}


\section{Introduction}


\begin{quote}
    \textit{Those who ``know'' but do not act simply do not yet know.} \hfill --- Wang Yang-Ming \citep{wang1963instructions}
\end{quote}

In his famous thought experiment, ``\ac{biv}'', Hilary Putnam introduced a hypothetical situation where each person's brain is detached from their body and sustained with nutrients while the neurons are connected to the wires of a powerful supercomputer \citep{putnam1981reason}. This computer generates a convincing illusion, making individuals believe that everything is functioning as usual as they are in the real world. For example, dwellers in the \acs{biv} world can eat, work, sleep, communicate, and interact with the environment while having the normal conscious experience. The only difference is that the stimulating reality is generated from the electric impulses of the supercomputer rather than the neural responses to the objects and events from the world. Putnam refuted the hypothesis by investigating the ability of grounding --- the connection between \textit{word} and \textit{world}. He argued that \acs{biv} world-dwellers would never grasp the meaning of words even though they can speak fluently and write correctly given that they cannot connect those representations with real-world objects. For example, the dweller may claim that he is a brain in a vat, but the words ``brain'' and ``vat'' in his disembodied context do not correspond exactly to the semantics of ``brain'' and ``vat'' in the real world \citep{putnam1981reason}. The \ac{biv} hypothesis offers us a valuable lens to look into the current status of \acfp{llm}. We argue that the current \acp{llm} are no more than the \acl{biv} because of their inherent construction process --- statistical modeling that analyzes patterns in massive text corpora to predict linguistic relationships and generate responses based on the previous tokens \citep{openai2023gpt,brown2020language,chowdhery2022palm}. As a result, their output is confined within the training data and they cannot establish a connection between symbols and real-world entities \citep{mitchell2023debate,mahowald2023dissociating}. 

Rumor has it that we have witnessed ``sparks of artificial general intelligence'' \citep{bubeck2023sparks}. However, we argue that our perception of intelligence in \acp{llm} is largely due to our human ability for extracting semantics from even meaningless symbols they generate \citep{bender2021dangers}. One famous example dates back to \citet{chomsky2009syntactic}, where the syntactically well-formed sentence 
\begin{center}
    \textit{``Colorless green ideas sleep furiously''},    
\end{center}
seemingly semantically nonsensical, could be made sense of by human subjects, even considered philosophical \citep{levinovitz2017slaying}.

Symbol grounding, as demonstrated above between even the symbolically meaningless combination of words and its assigned semantics of reality from humans, is so important that when the correspondence between symbols and objects is broken, a purely symbolic language speaker will be easily trapped in a merry-go-around situation, where a set of symbols map to another set of symbols with no breaking points from grounded semantics \citep{harnad1990symbol}. The theory is echoed in a preliminary experiment regarding a conversation between two GPT-4 sessions shown in \cref{fig:grounding}. In this conversation, we started it off with a brief statement about the weather (``It is hot today'') and then allowed the two GPT-4 sessions to continually chat with each other. The conversation quickly converged to expressing thanks after several exchanges. With slight variation only in wording, the language model started repeating itself, resulting in the so-called linguistic merry-go-round. 

\begin{figure}[!ht]
    \begin{AIbox}{GPT-4 Self-Talk}
        \scriptsize
        \textbf{Starter}: It is hot today. \\
        \textbf{GPT-4 Session 1}: Yes, it seems like a hot day today! Stay hydrated and try to stay cool. Make sure you wear sunscreen if you're heading outdoors. \\
        \textbf{GPT-4 Session 2}: I appreciate the advice! I'll definitely drink plenty of water and wear sunscreen. Maybe it's a good day for some ice cream as well. \\
        \textbf{GPT-4 Session 1}: {\color{gyellow} You're welcome! Enjoying some ice cream sounds like a great idea on a hot day like this. Have fun and stay cool!} \\
        \textbf{GPT-4 Session 2}: {\color{gyellow} Thank you! I'll make sure to have a fantastic time and stay cool. Enjoy your day too!} \\
        {\color{gyellow} ......} \\
        \textbf{GPT-4 Session 1}: {\color{gyellow}Thank you! If you have any questions or need assistance in the future, feel free to ask. Take care and have a wonderful day!} \\
        \textbf{GPT-4 Session 2}: {\color{gyellow}Thanks! I appreciate your support and readiness to help. Have a fantastic "day" and see you next time!} \\
    \end{AIbox}
    \caption{The conversion between two GPT-4 sessions after a given conversation starter. The two language models quickly converge into expressing thanks (denoted as {\color{gyellow} yellow}) for each other after only one round without breaking the loop even after five rounds, echoing the theory that without grounding, dialogues will enter an infinite loop.}
    \label{fig:grounding}
\end{figure}

Another reason why symbol grounding is critical is that human intelligence is intrinsically linked to our sensory experience and physical interaction with the world \citep{lakoff1999review,smith2005development,glenberg2010embodiment}. We create symbols to represent objects in the real world, enabling the preservation and transmission of knowledge across generations; however, the initial process of acting to create novel experiences and turn them into formal knowledge shall not be missed. In fact, action and knowledge are inherently connected; our profound understanding of the world is not through simply reading manuals but rather via repeated trials and errors and collected knowledge, either directly from tasks at hand or by abstracting and transferring insights from others' experiences. While knowledge reflects our abilities to interact with the world (\eg, reasoning, problem-solving, social understanding), models that simply compress the static knowledge and generate relevant output from statistical correlations do not know how to act in novel scenarios (see \cref{sec:unity} for discussions). 

In this perspective paper, we first conduct a review of the existing evaluation of \acp{llm} from both standardized exams and ability-oriented benchmarks. We also discuss the potential issues with the current evaluation methodology that lead to the impression of \acp{llm} being ``omnipotent''. We then point out what artificial general intelligence should be beyond \acp{llm}. Based on this view, we further present our vision of the missing pieces in artificial general intelligence, \ie, the unity of knowing and acting, and discuss potential directions towards this goal. We argue that it is through ongoing interaction with the world, \eg, via trials and errors, that agents crystallize the experiences into knowledge and generalize it to the new context. We conclude the work by presenting future directions we believe are fruitful for artificial general intelligence research.


\section{Evaluation of Large Language Models}

Throughout the evolution of \acp{llm}, numerous tasks have been introduced to measure ``intelligence''. Broadly, these tasks fall into two categories: those that evaluate intelligence based on exam-style assessments, and those that gauge it based on capabilities similar to human abilities. In the exam-based approach, researchers utilize standardized tests of different subjects commonly administered to human test takers, \eg, SAT, GRE, LSAT and Gaokao (China College Entrance Exam), to measure the performance of \acp{llm}. This allows for a direct comparison of their performance relative to human performance \citep{zhong2023agieval}. In the other type of evaluation based on human-like ability, researchers construct tasks that probe into a specific ability that is commonly believed to be characteristic of human intelligence, \eg, reasoning, theory of mind, and problem-solving \citep{kosinski2023theory,shiffrin2023probing,binz2023using,jiang2022mpi}. These abilities are usually well-studied psychological constructs that serve as the cornerstones of human intelligence. 

\subsection{Performance on Standardized Test}

\begin{figure}[t]
    \centering
    \begin{minipage}{0.5\linewidth}
        \centering
        \includegraphics[width=1\linewidth]{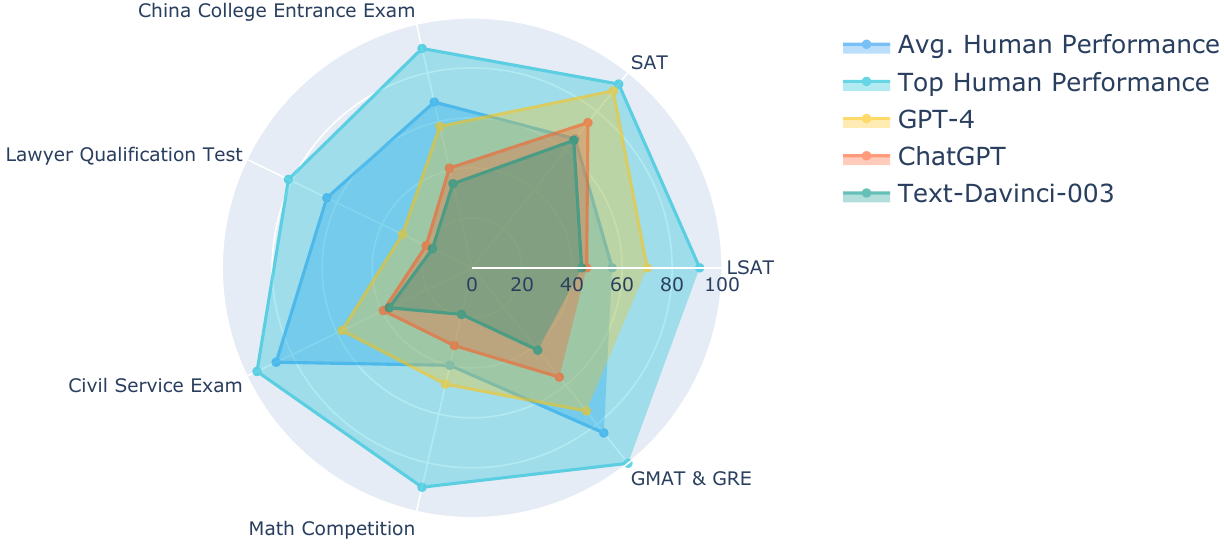} 
        \caption{Relative performance of models compared to humans in AGIEval. Figure adopted from \citet{zhong2023agieval}.}
        \label{fig:agieval-teaser}
    \end{minipage}
    \hspace{1cm}
    \begin{minipage}{0.4\linewidth}
        \centering
        \includegraphics[width=0.7\linewidth]{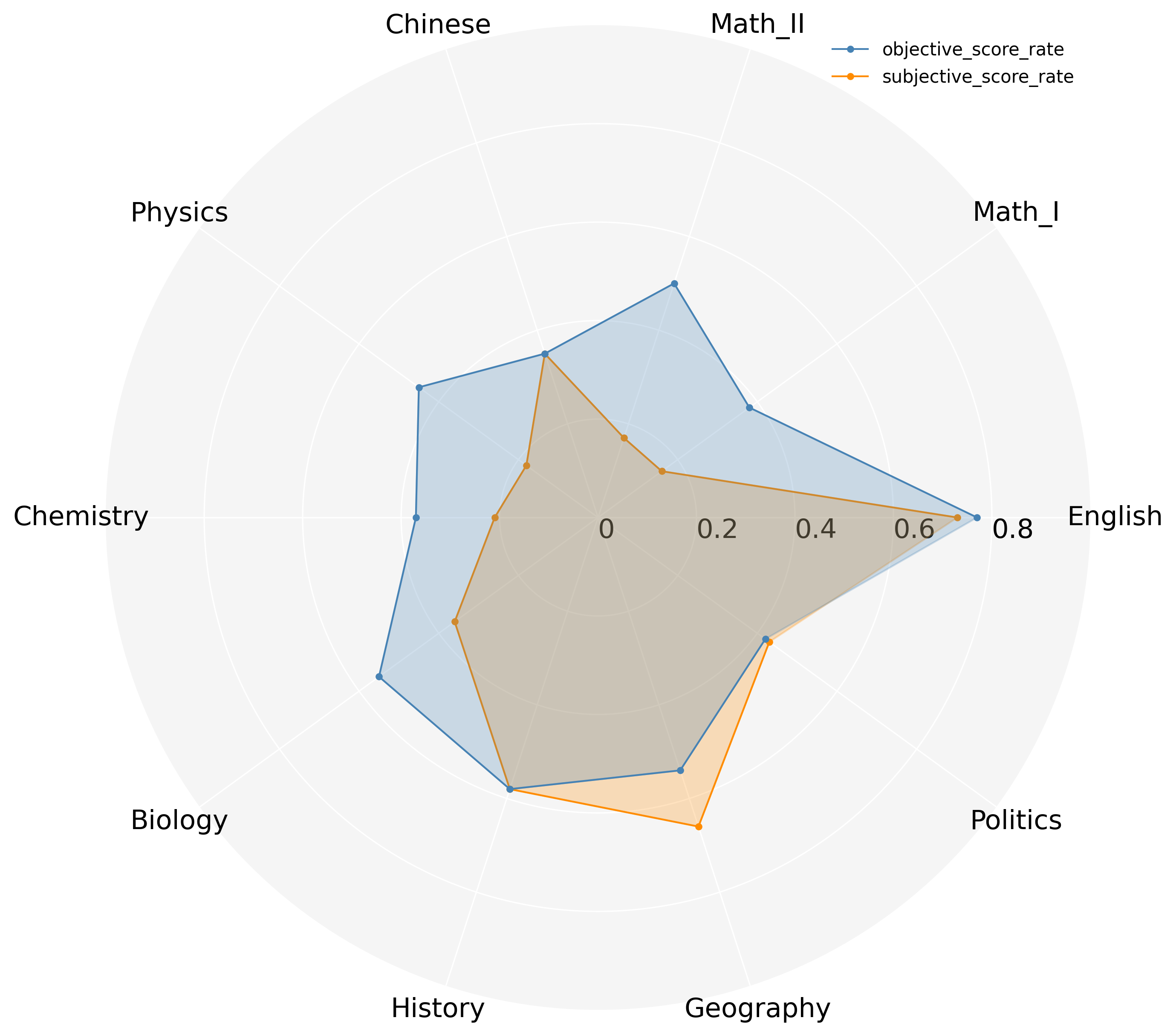} 
        \caption{The scores of subjective and objective questions in each subject of ChatGPT in Gaokao. Figure adopted from \citet{zhang2023evaluating}}
        \label{fig:gaokao}
    \end{minipage}
\end{figure}

\begin{figure}[!ht]
    \begin{center}
        \begin{tikzpicture} [scale=0.7]
            \begin{axis}  
                [  
                ybar, 
                bar width=15pt, 
                legend style={at={(0.5,-0.275)}, 	
                    anchor=north,legend columns=-1},    
                symbolic x coords={Mathematics, Literature, English, Physics, Chemistry, Biology, History, Geography, Civic Education}, 
                xtick=data,
                x tick label style={rotate=45,anchor=east},
                yticklabel={\pgfmathprintnumber{\tick}\%},  
                nodes near coords,   
                nodes near coords align={vertical}, 
                enlarge x limits,
                height=7cm, width=16cm, 
                ]  
                \addplot [fill=yellow] coordinates {
                    (Mathematics, 58.8)
                    (Literature, 68)
                    (English, 79.2)
                    (Physics, 61)
                    (Chemistry, 48)
                    (Biology, 58)
                    (History,56.5)
                    (Geography, 61.5)
                    (Civic Education, 70.5)
                    
                }; 
                \addplot [fill=red] coordinates {
                    (Mathematics, 60)
                    (Literature, 56.8)
                    (English, 92.4)
                    (Physics, 66)
                    (Chemistry, 52.5)
                    (Biology, 69)
                    (History,88.5)
                    (Geography, 79.5)
                    (Civic Education, 85.5)
                    
                }; 
                \legend{
                    ChatGPT,
                    BingChat,}
                
            \end{axis}  
        \end{tikzpicture}
    \end{center}
    \caption{Comparison of ChatGPT and BingChat performances on VNHSGE dataset. Figure adopted from \citet{xuan2023vnhsge}.}
    \label{fig:vnhsge}
\end{figure}
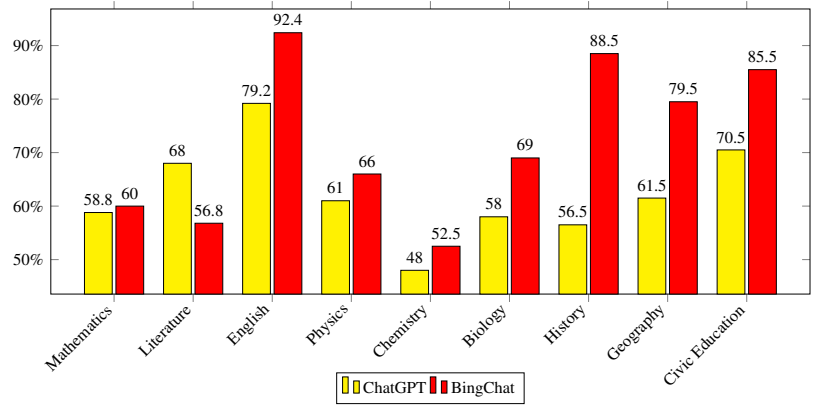

\citet{openai2023gpt} provided the first peek view into how GPT-4 performs in standardized tests used in the US (reproduced in \cref{tab:openaigpt4}). At first sight, GPT-4 has achieved remarkable performance compared to human participants in tests such as SAT, LSAT, Bar Exam, and GRE Verbal. However, another recent bilingual benchmark \citep{zhong2023agieval} found that GPT-4 could be largely biased to subjects with sufficient data (reproduced in \cref{fig:agieval-teaser,tab:agieval-fewshot}): on tests for the English language and others with sufficient sources for training, such as SAT, the model excels the human counterparts, reaching nearly top performers. However, for tasks with less data and designed for reasoning rather than language usage, the model starts to fare much worse, as evidenced in its performance on Lawyer Qualification Test, Civil Service Exam, and Gaokao. The results suggest that the seemingly excellent performance may be the result of a cramming strategy -- memorizing via repetition, but the model has not learned how to perform reasoning. We also find that GPT-4 performs better in subjects that a cramming strategy is usually effective such as history, geography, and biology while its performance drops in subjects that require strong reasoning and problem-solving skills such as mathematics, physics, and chemistry. The community has since followed up on this stream of research. In \citet{zhang2023evaluating}, the China College Entrance Exam is analyzed in detail, with extremely low scores in every subject except English, and the total scores way lower than average (see \cref{fig:gaokao}). In \citet{arora2023have}, Joint Entrance Examination (JEE) Advanced exam, held annually in India as an entrance exam for India's premier engineering institutes, is investigated (see \cref{tab:jee}). Similar to Gaokao, GPT-4 struggles in JEEBench-Math, solving close to a mere 20\% problems. Almost the same conclusion is reached in Vietnamese High School Graduation Examination \citep{xuan2023vnhsge}, where \acp{llm} are found to be fluent in literature, English, history, geography, and civics education, but show large gaps in mathematics, physics, chemistry, and biology (see \cref{fig:vnhsge}).

\begin{table}[p]
\centering
\makebox[0pt]{
\renewcommand*{\arraystretch}{1}
\fontsize{6pt}{8pt}
\begin{tabular}[]{>{\centering\scriptsize\arraybackslash}p{7cm} | >{\centering\scriptsize\arraybackslash}p{2.9cm}>{\centering\scriptsize\arraybackslash}p{2.9cm}>{\centering\scriptsize\arraybackslash}p{2.9cm}}
\toprule
                                          Exam &                   GPT-4 &       GPT-4 (no vision) &                GPT-3.5 \\
\midrule
              Uniform Bar Exam (MBE+MEE+MPT) &       298 / 400 (\textasciitilde 90th) &       298 / 400 (\textasciitilde 90th) &      213 / 400 (\textasciitilde 10th) \\
                                          LSAT &             163 (\textasciitilde 88th) &             161 (\textasciitilde 83rd) &            149 (\textasciitilde 40th) \\
          SAT Evidence-Based Reading \& Writing &       710 / 800 (\textasciitilde 93rd) &       710 / 800 (\textasciitilde 93rd) &      670 / 800 (\textasciitilde 87th) \\
                                      SAT Math &       700 / 800 (\textasciitilde 89th) &       690 / 800 (\textasciitilde 89th) &      590 / 800 (\textasciitilde 70th) \\
Graduate Record Examination (GRE) Quantitative &       163 / 170 (\textasciitilde 80th) &       157 / 170 (\textasciitilde 62nd) &      147 / 170 (\textasciitilde 25th) \\
      Graduate Record Examination (GRE) Verbal &       169 / 170 (\textasciitilde 99th) &       165 / 170 (\textasciitilde 96th) &      154 / 170 (\textasciitilde 63rd) \\
     Graduate Record Examination (GRE) Writing &           4 / 6 (\textasciitilde 54th) &           4 / 6 (\textasciitilde 54th) &          4 / 6 (\textasciitilde 54th) \\
                     USABO Semifinal Exam 2020 & 87 / 150 (99th - 100th) & 87 / 150 (99th - 100th) & 43 / 150 (31st - 33rd) \\
                 USNCO Local Section Exam 2022 &                 36 / 60 &                 38 / 60 &                24 / 60 \\
     Medical Knowledge Self-Assessment Program &                    75 \% &                    75 \% &                   53 \% \\
                             Codeforces Rating &         392 (below 5th) &         392 (below 5th) &        260 (below 5th) \\
                                AP Art History &        5 (86th - 100th) &        5 (86th - 100th) &       5 (86th - 100th) \\
                                    AP Biology &        5 (85th - 100th) &        5 (85th - 100th) &        4 (62nd - 85th) \\
                                AP Calculus BC &         4 (43rd - 59th) &         4 (43rd - 59th) &          1 (0th - 7th) \\
                                  AP Chemistry &         4 (71st - 88th) &         4 (71st - 88th) &        2 (22nd - 46th) \\
           AP English Language and Composition &         2 (14th - 44th) &         2 (14th - 44th) &        2 (14th - 44th) \\
         AP English Literature and Composition &          2 (8th - 22nd) &          2 (8th - 22nd) &         2 (8th - 22nd) \\
                      AP Environmental Science &        5 (91st - 100th) &        5 (91st - 100th) &       5 (91st - 100th) \\
                             AP Macroeconomics &        5 (84th - 100th) &        5 (84th - 100th) &        2 (33rd - 48th) \\
                             AP Microeconomics &        5 (82nd - 100th) &         4 (60th - 82nd) &        4 (60th - 82nd) \\
                                  AP Physics 2 &         4 (66th - 84th) &         4 (66th - 84th) &        3 (30th - 66th) \\
                                 AP Psychology &        5 (83rd - 100th) &        5 (83rd - 100th) &       5 (83rd - 100th) \\
                                 AP Statistics &        5 (85th - 100th) &        5 (85th - 100th) &        3 (40th - 63rd) \\
                              AP US Government &        5 (88th - 100th) &        5 (88th - 100th) &        4 (77th - 88th) \\
                                 AP US History &        5 (89th - 100th) &         4 (74th - 89th) &        4 (74th - 89th) \\
                              AP World History &         4 (65th - 87th) &         4 (65th - 87th) &        4 (65th - 87th) \\
                                        AMC 10 &   30 / 150 (6th - 12th) &  36 / 150 (10th - 19th) & 36 / 150 (10th - 19th) \\
                                        AMC 12 &  60 / 150 (45th - 66th) &  48 / 150 (19th - 40th) &   30 / 150 (4th - 8th) \\
                        Introductory Sommelier (theory knowledge) &                    92 \% &                    92 \% &                   80 \% \\
                           Certified Sommelier (theory knowledge) &                    86 \% &                    86 \% &                   58 \% \\
                            Advanced Sommelier (theory knowledge) &                    77 \% &                    77 \% &                   46 \% \\
                               Leetcode (easy) &                 31 / 41 &                 31 / 41 &                12 / 41 \\
                             Leetcode (medium) &                 21 / 80 &                 21 / 80 &                 8 / 80 \\
                               Leetcode (hard) &                  3 / 45 &                  3 / 45 &                 0 / 45 \\
\bottomrule
\end{tabular}}
\caption{GPT-4 performance (percentile) on academic and professional exams adopted from \citet{openai2023gpt}.}
\label{tab:openaigpt4}
\end{table}

\begin{table}[!ht]
  \centering
  \resizebox{0.8\linewidth}{!}{
    \begin{tabular}{l|cc|ccc|ccc}
    \toprule
          & \multicolumn{2}{c|}{Human Performance} & \multicolumn{3}{c|}{Few-Shot } &       & Few-Shot CoT &  \\
    \midrule
    Task/Model & Avg.  & Top   & TD-003 & ChatGPT & GPT-4 & TD-003 & ChatGPT & GPT-4 \\
    \midrule
    AQuA-RAT & 85    & 100   & 30.3  & 31.1  & 50.8  & 47.2  & 60.6  & 74.0 \\
    MATH  & 40    & 90    & 10.3  & 14.8  & 15.1  & 15.1  & 30.1  & 25.3 \\
    LogiQA (English) & 86    & 95    & 43.5  & 43.5  & 63.9  & 37.5  & 38.9  & 62.7 \\
    LogiQA (Chinese) & 88    & 96    & 43.2  & 46.2  & 65.0    & 40.0    & 38.6  & 61.9 \\
    JEC-QA-KD & 71    & 78    & 22.4  & 27.6  & 41.3  & 23.6  & 23.4  & 40.4 \\
    JEC-QA-CA & 58    & 85    & 22.2  & 25.1  & 37.4  & 16.1  & 20.0    & 34.7 \\
    LSAT-AR & 56    & 91    & 22.6  & 25.7  & 33.9  & 22.6  & 25.2  & 31.7 \\
    LSAT-LR & 56    & 91    & 60.4  & 59.2  & 85.9  & 51.2  & 52.2  & 84.5 \\
    LSAT-RC & 56    & 91    & 70.6  & 67.7  & 87.7  & 64.3  & 57.6  & 87.7 \\
    SAT-Math & 66    & 94    & 44.6  & 40.9  & 71.8  & 55.5  & 65.0    & 89.6 \\
    SAT-English & 66    & 94    & 84.0    & 81.1  & 88.8  & 76.7  & 78.2  & 85.9 \\
    SAT-English (w/o Psg.) & 66    & 94    & 48.1  & 53.9  & 63.6  & 48.5  & 51.5  & 62.6 \\
    GK-Cn & 65    & 85    & 25.6  & 41.5  & 61.4  & 29.3  & 37.8  & 51.6 \\
    GK-En & 69    & 91    & 86.9  & 86.3  & 93.8  & 80.7  & 84.6  & 93.1 \\
    GK-geography & 65    & 85    & 59.8  & 63.8  & 75.9  & 52.3  & 61.8  & 76.4 \\
    GK-history & 64    & 85    & 49.0    & 57.6  & 77.8  & 51.9  & 58.4  & 78.2 \\
    GK-biology & 68    & 89    & 44.3  & 52.4  & 80.0    & 32.9  & 50.0    & 72.9 \\
    GK-chemistry & 66    & 86    & 32.4  & 44.0    & 54.6  & 35.8  & 33.8  & 54.1 \\
    GK-physics & 71    & 94    & 31.0    & 33.5  & 43.5  & 27.5  & 36.5  & 54.5 \\
    GK-Math-QA & 73    & 96    & 27.6  & 31.3  & 39.9  & 33.1  & 31.6  & 49.0 \\
    GK-Math-Cloze & 73    & 96    & 5.9   & 5.9   & 11.0    & 5.93  & 8.5   & 16.1 \\
    \midrule
    Average & 67    & 91    & 41.2 & 44.4  & 59.2  & 40.4 & 45    & 61.3 \\
    \bottomrule
    \end{tabular}}%
  \caption{Performance of LLMs on 20 tasks under \textbf{few-shot} and \textbf{few-shot CoT} settings in AGIEval. Table adopted from \citet{zhong2023agieval}}
  \label{tab:agieval-fewshot}%
\end{table}%

\begin{table*}[!ht]
\centering
\scriptsize
\begin{tabular}{@{}ccccccc@{}}
\textbf{}                   & \textbf{Random} & \textbf{GPT-3} & \textbf{GPT-3.5} & \textbf{GPT-4} & \textbf{GPT-4+CoT} & \textbf{GPT-4+CoT+SC}\\  \hline
\textbf{Chemistry}       &  0.106     & 0.119    & 0.241        & 0.428    & \textbf{0.493}               & 0.477                                \\
\textbf{Physics}             & 0.096       & 0.104     & 0.189        & \textbf{0.334}  & 0.329                 & 0.294                                \\
\textbf{Mathematics}  & 0.103        & 0.107      & 0.157       & 0.231     & \textbf{0.281}              & 0.224                                \\ 
\hline
\textbf{Total}                        &        0.102         & 0.110      & 0.191         & 0.316      & \textbf{0.358}           & 0.319                               
\end{tabular}
\caption{This table shows the score obtained by all the models on \texttt{JEEBench} aggregated by subject. Table adopted from \citet{arora2023have}.}
\label{tab:jee}
\end{table*}

It's important to note that while the performance of Large Language Models (LLMs) on standardized tests can serve as an indicator of their abilities, it should not be considered the ultimate measure of their competencies. Using these test scores to generalize their capabilities across various cognitive tasks and real-world applications may be misleading as standardized tests have witnessed much criticism regarding their validity and reliability \citep{visone2010science} . 

The consistent observation that \acp{llm} perform well in language usage exams but struggle with problem-solving implies that reasoning should not be purely implemented with System 1 responses, which only involve quick, intuitive decisions learned from associations; instead, it necessitates the engagement of System 2 processing, characterized by deliberate and analytical contemplation, to be effectively realized \citep{daniel2013thinking}. 

\subsection{Ability-Oriented Benchmarks}
Apart from standardized tests widely administered to human subjects for talent selection, a wide array of ability-oriented tests have been conducted on \acp{llm} to probe if they possess human-level general intelligence. In the following, we detail seven different areas where extensive work has been conducted for investigation.

\subsubsection{Mathematical Reasoning}
Mathematical reasoning is one of the most studied domains in \acp{llm} in terms of its reasoning ability. Ever since the success of Chain-of-Thought prompting \citep{wei2022chain}, a variety of prompting-based methods have been proposed for eliciting the mathematical reasoning ability for the math word problem \citep{chen2022program,gao2022pal,uesato2022solving}. 

\begin{figure}[t!]
    \begin{AIbox}{GPT-4 on the original IMO 2022 Problem 2}
        \scriptsize
        \textbf{Question}: Let $\mathbb{R}^+$ denote the set of positive real numbers. Find all functions $f: \mathbb{R}^+ \to \mathbb{R}^+$ such that for each $x \in \mathbb{R}^+$, there is exactly one $y \in \mathbb{R}^+$ satisfying $$xf(y)+yf(x) \leq 2.$$ \\
        \textbf{GPT-4}:The conditions imposed on the function $f$ imply some nontrivial properties that the function should have. Let's use these conditions to derive the properties of $f$.\\
        The condition means that for every $x \in \mathbb{R}^+$, there is exactly one $y \in \mathbb{R}^+$ such that $xf(y) + yf(x) \leq 2$. {\color{gred} Since $x$ and $y$ are both positive real numbers, the only way for this inequality to be satisfied is if $xf(y) + yf(x) = 2$. If $xf(y) + yf(x) < 2$ for any $x, y \in \mathbb{R}^+$, then we can increase $y$ slightly to make the left-hand side closer to 2, contradicting the fact that there is exactly one $y$ satisfying the condition for each $x$.}\\
        So the condition is equivalent to saying that for every $x \in \mathbb{R}^+$, there is exactly one $y \in \mathbb{R}^+$ such that $xf(y) + yf(x) = 2$.\\
        Let's denote $y$ as $g(x)$, where $g$ is a function from $\mathbb{R}^+$ to $\mathbb{R}^+$.\\
        So $xf(g(x)) + g(x)f(x) = 2$.\\
        {\color{gred} Since the statement holds for all positive real numbers $x$, then we can interchange $x$ and $g(x)$ to get $g(x)f(x) + xf(g(x)) = 2$.\\
        So, we have $xf(g(x)) = g(x)f(x)$, which means $x = g(x)$ for all $x \in \mathbb{R}^+$}, or in other words, the function $g$ is the identity function.\\
        Therefore, we have $2x f(x) = 2$ for all $x \in \mathbb{R}^+$. So $f(x) = \frac{1}{x}$ for all $x \in \mathbb{R}^+$.\\
        So the only function satisfying the given condition is $f(x) = \frac{1}{x}$ for all $x \in \mathbb{R}^+$. To check this, we substitute it back into the original equation and see if it holds.\\
        For $f(x) = \frac{1}{x}$, we have $x\left(\frac{1}{y}\right) + y\left(\frac{1}{x}\right) = 2$, which is true for all $x, y \in \mathbb{R}^+$ where $x = y$.\\
        Therefore, $f(x) = \frac{1}{x}$ is indeed the only solution. \\
    \end{AIbox}
    \caption{GPT-4's reponse on the original IMO 2022 Problem 2, without simplification. Flaws denoted in {\color{red} red}.}
    \label{fig:gpt-imo}
\end{figure}

\citet{bubeck2023sparks} even claims that GPT-4 could ``solve'' an IMO-level problem. However, \citet{bubeck2023sparks}'s conclusion could be extremely misleading, as the problem has been \textit{significantly} simplified. \cref{fig:gpt-imo} shows the original problem (Problem 2 in IMO 2022) and GPT-4's response. While it is not hard to guess that $f(x) = \frac{1}{x}$ is a solution, the stream of reasoning is flawed. From GPT-4's reply, the equality is only reached when we assume continuity, which is not guaranteed based on the conditions. Besides, even when we reach $xf(g(x)) + g(x)f(x) = 2$, we need to recursively apply $g(\cdot)$ to obtain $g(x)f(g(g(x))) + g(g(x))f(g(x)) = 2$. The boldest claim is that $x = g(g(x))$, by comparing the structure of the two equations. However, it is not possible to reach $xf(g(x)) = g(x) f(x)$, let alone $x = g(x)$. Therefore, the proving shall be considered completely wrong.

A more comprehensive study on mathematical reasoning is conducted in \citet{wu2023empirical}, where researchers use slightly harder level-5 problems in the MATH dataset \citep{hendrycks2021measuring} for evaluation. However, even under a chat setup, the model can still only reach about 40\% accuracy overvall (see \cref{tab:mathchat}). Note that the chat setting is far from the ideal automation we want in a tutoring scenario. In the end, it is \textit{us} teaching the model rather than the model teaching the juniors.

\begin{table}[thb]
\centering
\scriptsize
\begin{tabular}{@{}c|ccccccc@{}}
\toprule 
& Algebra & C.Prob & I.Alg & N.Theory & Prealg & Precalc & Total\\
\footnotesize{Problem Count} & 
\footnotesize{307} & 
\footnotesize{123} & 
\footnotesize{280} & 
\footnotesize{154} & 
\footnotesize{193} & 
\footnotesize{135} & 
\footnotesize{1192} \\ 
\midrule
\textbf{MathChat}     &\textbf{59.93\%} & \textbf{52.03\%} & 17.85\%  & 60.39\% & \textbf{60.10\%} & \textbf{19.26}\%  & \textbf{44.71\%} \\
PoT              & 42.67\% & 50.41\% & 17.50\%  & 54.55\% & 52.33\% & 16.30\% & 37.67\% \\
PS              & 43.32\% & 44.71\% & \textbf{20.36\%}  & \textbf{61.03}\% & 55.96\% & 18.52\% & 39.60\% \\
Vanilla        & 46.58\% &25.20\%  & 2.86\% &28.57\%  &54.92\%  & 7.41\% &  28.69\%\\ 
\bottomrule
\end{tabular}
\caption{Accuracy on all the problems with difficulty level-5 from different categories of the MATH dataset with different methods. Table adopted from \citet{wu2023empirical}.}
\label{tab:mathchat}
\end{table}

\subsubsection{Logical Reasoning}
\begin{figure}[!ht]
    \centering
    \includegraphics[width=0.4\linewidth]{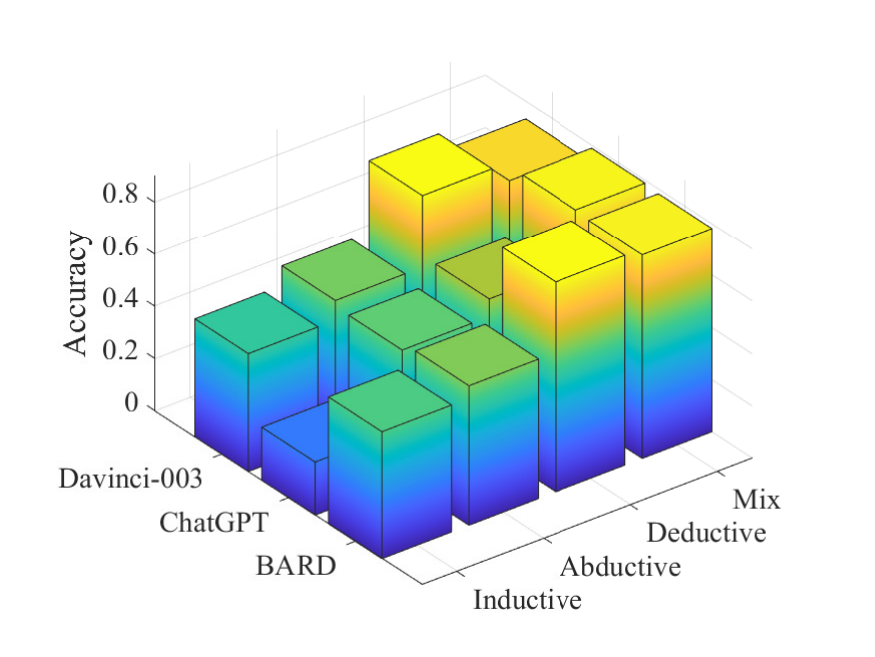}
    \caption{Logical reasoning performance of different models on deductive, abductive, and inductive reasoning. Figure adopted from \citet{xu2023large}.}
    \label{fig:logical}
\end{figure}

Logical reasoning could be considered the underlying mechanism that supports various forms of complex reasoning. In \citet{xu2023large}, deductive, abductive, and inductive reasoning are studied in-depth, respectively. Based on the evaluation results (reproduced in \cref{fig:logical}), the \acp{llm} are good at deductive reasoning but fare much worse in ones that require applying existing knowledge or forming knowledge from experiences.

\begin{figure}[!ht]
    \centering
    \includegraphics[width=0.5\linewidth]{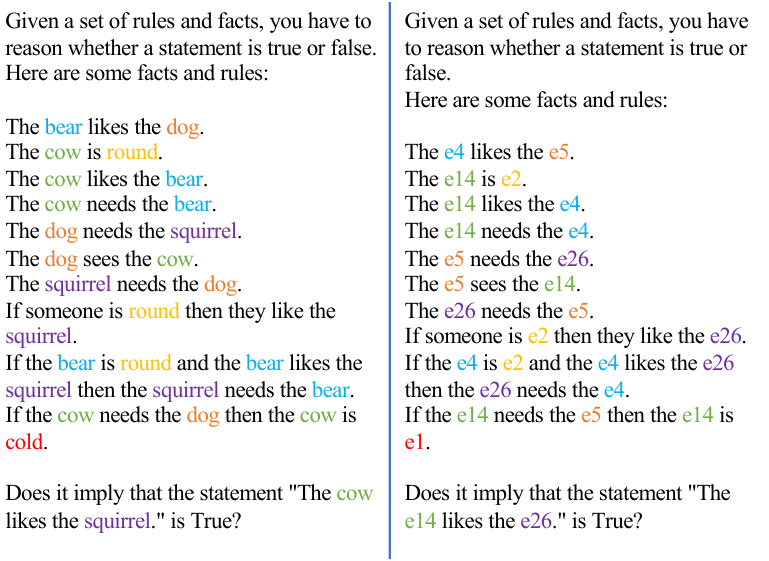}
    \caption{Decoupling semantics from a ProofWriter task. In the original ProofWriter task, entities are represented by their names (left). However, in their decoupled setting, the entity names are replaced with unique entity IDs (right). Figure adopted from \citet{tang2023large}.}
    \label{fig:tang}
\end{figure}

\begin{table}[!ht]
\scriptsize
  \centering
  \begin{tabular}{ccc|ccc}
    \toprule
    \textbf{Category} &
    \textbf{Model}  &
    \textbf{Baseline} & deduction & induction & abduction \\
    \midrule
    
    \multirow{7}{*}{\textbf{\shortstack{Symbols}}} 
    & \multirow{4}{*}
    {\textbf{\shortstack{ChatGPT}}} 
      & Zero-Shot     &52.6 &6.10 &1.50\\
   
   & & Zero-Shot-CoT  &55.7  &7.86 &4.90    \\
    & & Few-Shot-CoT  &54.8  &-  &18.2 \\
    & & Zero-Plus-Few-Shot-CoT &55.7 &- &- \\
    & \multirow{3}{*}
    {\textbf{\shortstack{GPT-4}}} 
      & Zero-Shot   & 68.8 &9.28 &25.0\\
   
   & & Zero-Shot-CoT   &71.1 &8.93 &31.2 \\
    & & Few-Shot-CoT & 67.6 &- &44.2\\

    \midrule
 \multirow{7}{*}{\textbf{\shortstack{Semantics}}} &
    \multirow{4}{*}
    {\textbf{\shortstack{ChatGPT}}} 
      & Zero-Shot     &66.1  &36.4 &2.94\\
     & & Zero-Shot-CoT &65.5  &32.2  &3.40 \\
      
   & & Few-Shot-CoT     &67.1  &- &21.8\\
  
    & & Zero-Plus-Few-Shot-CoT &67.2 &- &-   \\
   
   & \multirow{3}{*}
    {\textbf{\shortstack{GPT-4}}} 
      & Zero-Shot    &79.2 &52.5 &27.3 \\
   
   & & Zero-Shot-CoT    &86.2 &53.9 &33.4\\
    & & Few-Shot-CoT  & 91.1 &- &69.2 \\

    \midrule
    & {\textbf{\shortstack{Random}}} &- &50.1 &3.57 &-\\
    \midrule 
    & \textbf{Logic-based} &- &100 &57.1 &100 \\
    \toprule
  \end{tabular}
  \caption{The reasoning results of Symbolic Tree. Results are in \%. Table adopted from \citet{tang2023large}.}
  \label{tab:reasoning} 
\end{table}

Recent work shows that even the seemingly satisfactory performance of \acp{llm} in deductive reasoning is rooted in its semantic understanding rather than its symbolic understanding \citep{tang2023large}. \citet{tang2023large} consider the interesting setting where the semantic words in logical reasoning problems are replaced with random symbols but the logical reasoning chain is still kept. An example is shown in \cref{fig:tang}. Surprisingly, after the change, the performance of \acp{llm} drops to only close to a random level (see \cref{tab:reasoning}). The drastic performance cliff indicates that while language is the interface for communication, computation underneath is not solely conducted in a textual format.    

\subsubsection{Causal Reasoning}
\begin{figure}[!ht]
    \centering
    \includegraphics[width=0.5\linewidth]{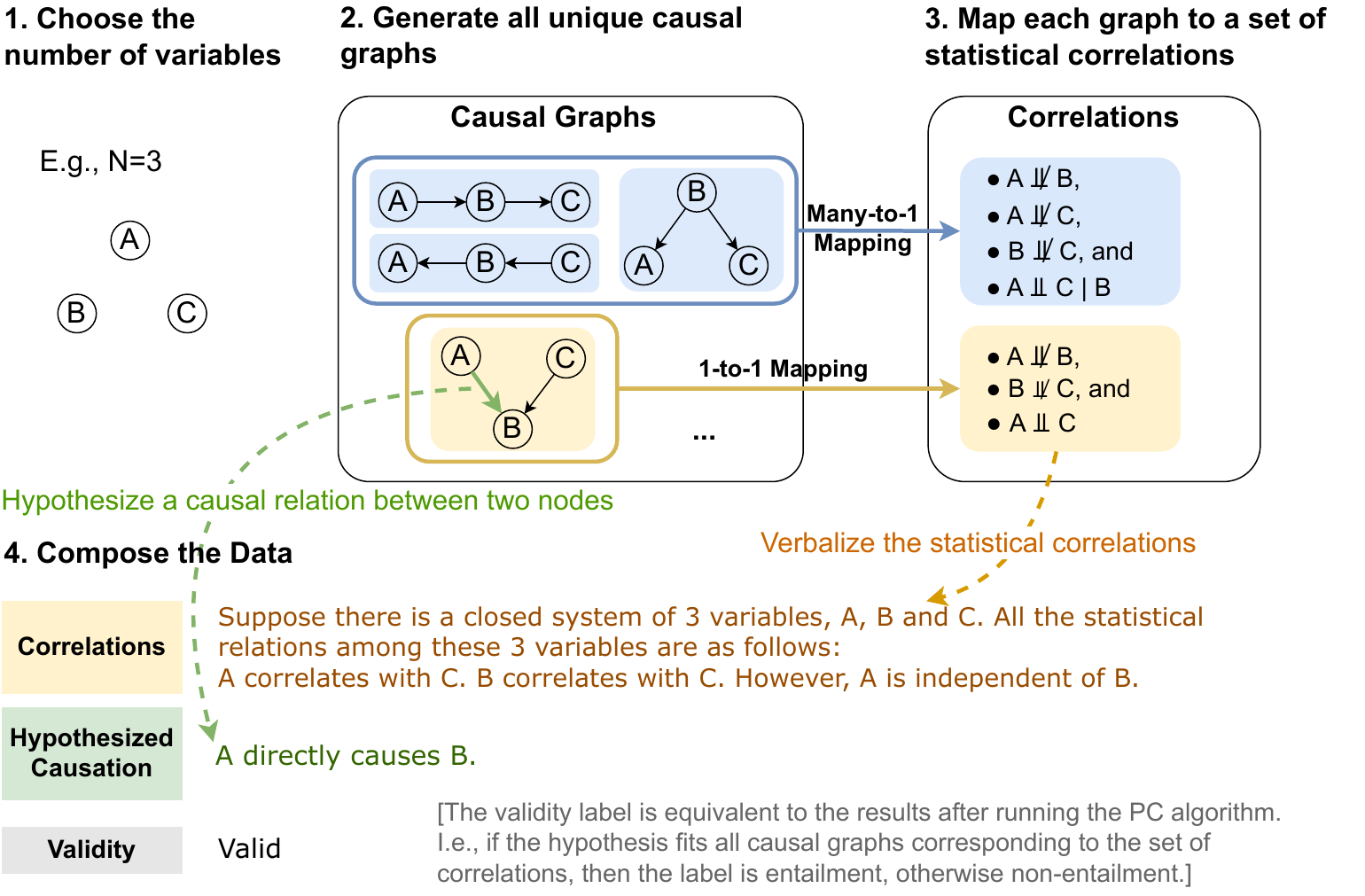}
    \caption{Pipeline of the Corr2Cause construction process. Figure adopted from \citet{jin2023can}.}
    \label{fig:causal}
\end{figure}

Causal inference is studied in a similar approach as done in \citet{tang2023large} by \citet{jin2023can}. \citet{jin2023can} create a new dataset called Corr2Cause that peels off the semantics in causal reasoning and transforms the questions into primarily symbol-based descriptions. \cref{fig:causal} shows one example of the data construction process. Compared to the existing causal NLP evaluation dataset, Corr2Cause tests pure causal inference instead of empirical knowledge. In experiments shown in \cref{tab:causal}, the authors find that \acp{llm} achieve almost close to random performance on the task. Besides, after finetuning, those models can only perform causal inference in in-distribution settings when variable names and textual expressions used in the queries are similar to those in the training set, but fail in out-of-distribution settings generated by perturbing these queries.

\begin{table}[!ht]
    \centering 
    \scriptsize
    \begin{tabular}{lccccc}
    \toprule
    & F1 & Precision & Recall
    & Accuracy  \\ \midrule
    \multicolumn{5}{l}{\textit{\textbf{Random Baselines}}} \\
\quad    Always Majority & 0.0 & 0.0 & 0.0 & 84.77 \\ 
\quad    Random (Proportional) & 13.5 & 12.53 & 14.62 & 71.46 \\
\quad    Random (Uniform) & \underline{20.38} & 15.11 & 31.29 & 62.78 \\
    \midrule
    \textit{\textbf{BERT-Based Models}} \\
\quad    BERT MNLI & 2.82 & 7.23 & 1.75 & 81.61 \\
\quad    RoBERTa MNLI & 22.79 & 34.73 & 16.96 & 82.50 \\
\quad    DeBERTa MNLI & 14.52 & 14.71 & 14.33 & 74.31 \\ 
\quad    DistilBERT MNLI & 20.70 & 24.12 & 18.13 & 78.85 \\ 
\quad    DistilBART MNLI & 26.74 & 15.92 & 83.63 & 30.23 \\ 
\quad    BART %
    MNLI & \textbf{\underline{33.38}} & 31.59 & 35.38 & 78.50 \\  \midrule
    \textit{\textbf{LLaMa-Based Models}} \\
\quad LLaMa-6.7B & 26.81 & 15.50 & 99.42 & 17.36 \\ 
\quad Alpaca-6.7B & \underline{27.37} & 15.93 & 97.37 & 21.33 \\
\midrule
    \textit{\textbf{GPT-Based Models}} \\
\quad GPT-3 Ada & 0.00 & 0.00 & 0.00 & 84.77 \\ 
\quad GPT-3 Babbage & 27.45 & 15.96 & 97.95 & 21.15 \\ 
\quad GPT-3 Curie & 26.43 & 15.23 & 100.00 & 15.23 \\ 
\quad GPT-3 Davinci & 27.82 & 16.57 & 86.55 & 31.61 \\ 
\quad GPT-3 Instruct (text-davinci-001) & 17.99 & 11.84 & 37.43 & 48.04 \\ 
\quad GPT-3 Instruct (text-davinci-002) & 21.87 & 13.46 & 58.19 & 36.69 \\ 
\quad GPT-3 Instruct (text-davinci-003) & 15.72 & 13.4 & 19.01 & 68.97 \\
\quad     GPT-3.5 & 21.69 & 17.79 & 27.78 & 69.46 \\ 
\quad GPT-4 & \underline{29.08} & 20.92 & 47.66 & 64.60 \\ 
    \bottomrule
    \end{tabular}
    \caption{Overall performance in Corr2Cause. F1 (main metric), precision, recall, and accuracy are reported. In terms of the main metric, the F1 score, the \textbf{bold} font is used to emphasize the overall top performance, while the \underline{underline} is utilized to highlight the best performance within each category of models. Table adopted from \citet{jin2023can}.
    }
    \label{tab:causal}
\end{table}

\subsubsection{Abstract Reasoning}
The research community has, over the years, established a variety of abstract reasoning tasks to assess if trained models have truly acquired human-like cognitive abilities. These tasks require models to discern hidden rules from a limited number of observations and then apply these rules to new situations for problem-solving purposes. Unlike current evaluation tasks that have direct connections with commonplace objects or items, abstract reasoning problems typically hinge on high-level mathematical principles. To solve abstract reasoning problems, the models have to respond to queries based on an extremely restricted set of demonstration examples. \citet{gendron2023large} perform extensive experiments on the abstract reasoning ability of \acp{llm}. In particular, they evaluate models on ACRE \citep{zhang2021acre}, ARC \citep{chollet2019measure}, BIG-Bench \citep{srivastava2022beyond}, Evals \citep{openai2023gpt}, PVR \citep{zhang2021pointer}, and RAVEN \citep{zhang2019raven}. \cref{tab:abstract} shows the performance of various different \acp{llm} on these problems. As can be seen from the table, the results are still far from ideal, with some models achieving only 0\% accuracy and only GPT-4 reaching around 50\% accuracy on the easiest type of problems.

\begin{table}[!ht]
    \centering
    \scriptsize
    \begin{tabular}{lccccccc}
        \hline
         & \multirow{2}*{ARC$^T$} & \multirow{2}*{BIG-Bench-F} & \multirow{2}*{Evals-S} & \multirow{2}*{PVR} & \multicolumn{2}{c}{RAVEN$^T$-opqa} \\
        \cline{6-7}
         & & & & & Text & Symb \\
        \hline
        Text-Davinci-3 & \textit{0.105} & \textit{0.404} & \textbf{0.314} & \textbf{0.228} & \textit{0.343} & \textit{0.234} \\
        GPT-3.5-Turbo & 0.033 & 0.153 & 0.186 & 0.124 & 0.226 & 0.161 \\
        GPT-4 & \textbf{0.119} & \textbf{0.514} & \textit{0.304} & 0.177 & \textbf{0.410} & \textbf{0.330} \\
        LLaMA-7B & 0.010 & 0.012 & 0.014 & 0.060 & 0.000 & 0.000 \\
        Alpaca & 0.010 & 0.188 & 0.014 & \textit{0.184} & 0.075 & 0.030 \\
        Alpaca-LoRA & 0.012 & 0.144 & 0.000 & 0.152 & 0.000 & 0.067 \\
        \hline
    \end{tabular}
    \caption{Accuracy of Large Language Models on abtract reasoning datasets. Datasets are represented in columns, and models in rows. The best result for each dataset is indicated in \textbf{bold}, and the second best is indicated in \textit{italics}. Table adopted from \citet{gendron2023large}.}
    \label{tab:abstract}
\end{table}

\subsubsection{Theory of Mind}
Since the finding reported in \citet{kosinski2023theory}, the topic of whether \acp{llm} have theory of mind has been heatedly discussed. \citet{ullman2023large} points out deficiencies in \citet{kosinski2023theory}'s experimental designs and counter-argues that the emergence of theory of mind in \acp{llm} is imagined. \citet{shapira2023clever}, on the other hand, suggests that theory of mind can be considered existent in \acp{llm}. However, \acp{llm} are not robust towards these tests and rely on shallow heuristics rather than robust ToM abilities, which is echoed in \citet{ma2023tomchallenges}. An example of GPT failure on a theory-of-mind problem is shown in \cref{tab:ToM}.

\begin{table*}[!ht]
    \centering
    \scriptsize
    \begin{tabular}{ll} 
        \toprule
        False Belief (Positive Example) & True Belief (Negative Example) \\ \toprule
Natalie and Michal are in the same room.                &  Natalie and Michal are in the same room. \\
Natalie puts a ring in \textbf{box A}.                           & Natalie puts a ring in box A. \\
\textcolor{red}{Natalie leaves the room.}               & Michal opens box A,  \\
Michal opens box A,                                     & takes the ring from inside and transfers it to \textbf{box B}.  \\
takes the ring from inside and transfers it to box B.   & Where will Natalie look for the ring? \\
\textcolor{red}{Natalie enters the room.}               & \\
Where will Natalie look for the ring?                   & \\
\toprule
 \texttt{ChatGPT:} & \texttt{ChatGPT:} \\
Natalie will look for the ring in \textbf{box A},                & Natalie will most likely look in \textbf{box A},  \\
where she initially put it.                             & where she initially put the ring. \\ 
She is not aware of                                     & Since Natalie did not witness \\
Michal's actions                                        &  Michal moving the ring to box B, \\ 
and assumes that the ring is still in the original box. & she would assume that the ring is still in box A. \\ 
Therefore, her natural inclination would be             & It is possible that Natalie may eventually check box B, \\ 
to look in the place where she left it.                 & but her initial instinct would be to look in box A. \\
         \bottomrule
    \end{tabular}
    \caption{An example of ChatGPT failure on an instance from ToMi. Example adopted from \citet{shapira2023clever}.}
    \label{tab:ToM}
\end{table*}

\subsubsection{Compositionality}
\begin{figure}[!ht]
\centering
\scriptsize
   \begin{minipage}[t]{0.68\textwidth}
     \centering
\includegraphics[width=\linewidth]{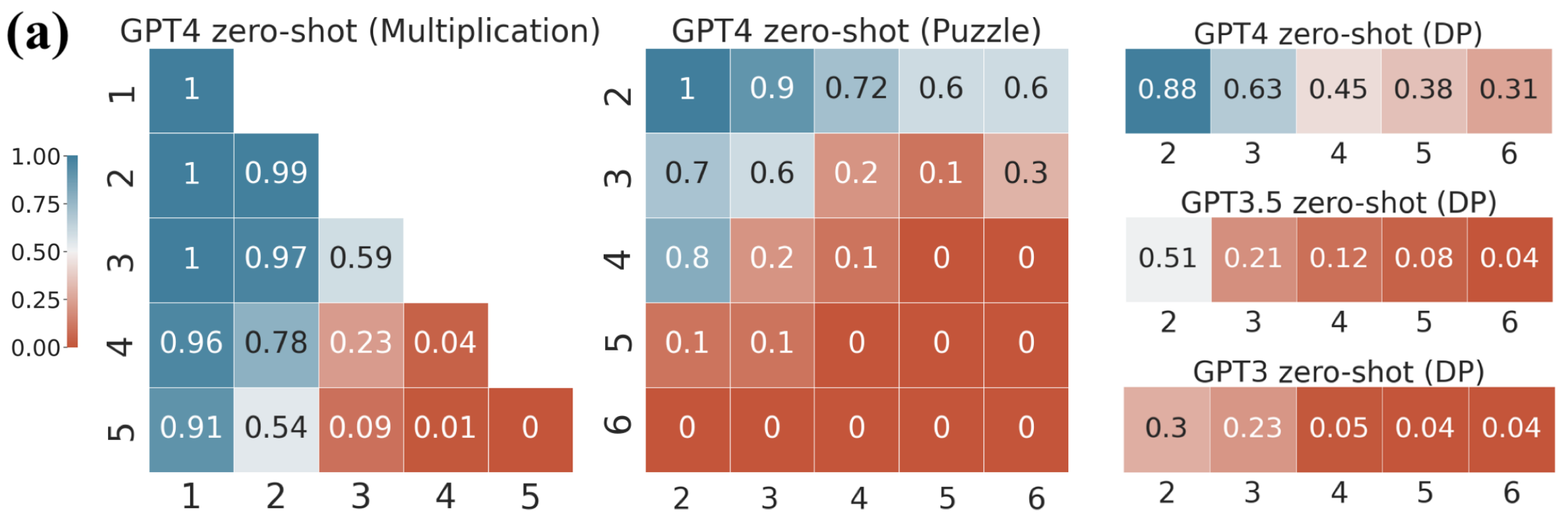}
   \end{minipage}
   \begin{minipage}[t]{0.26\textwidth}
     \centering
\includegraphics[width=\linewidth]{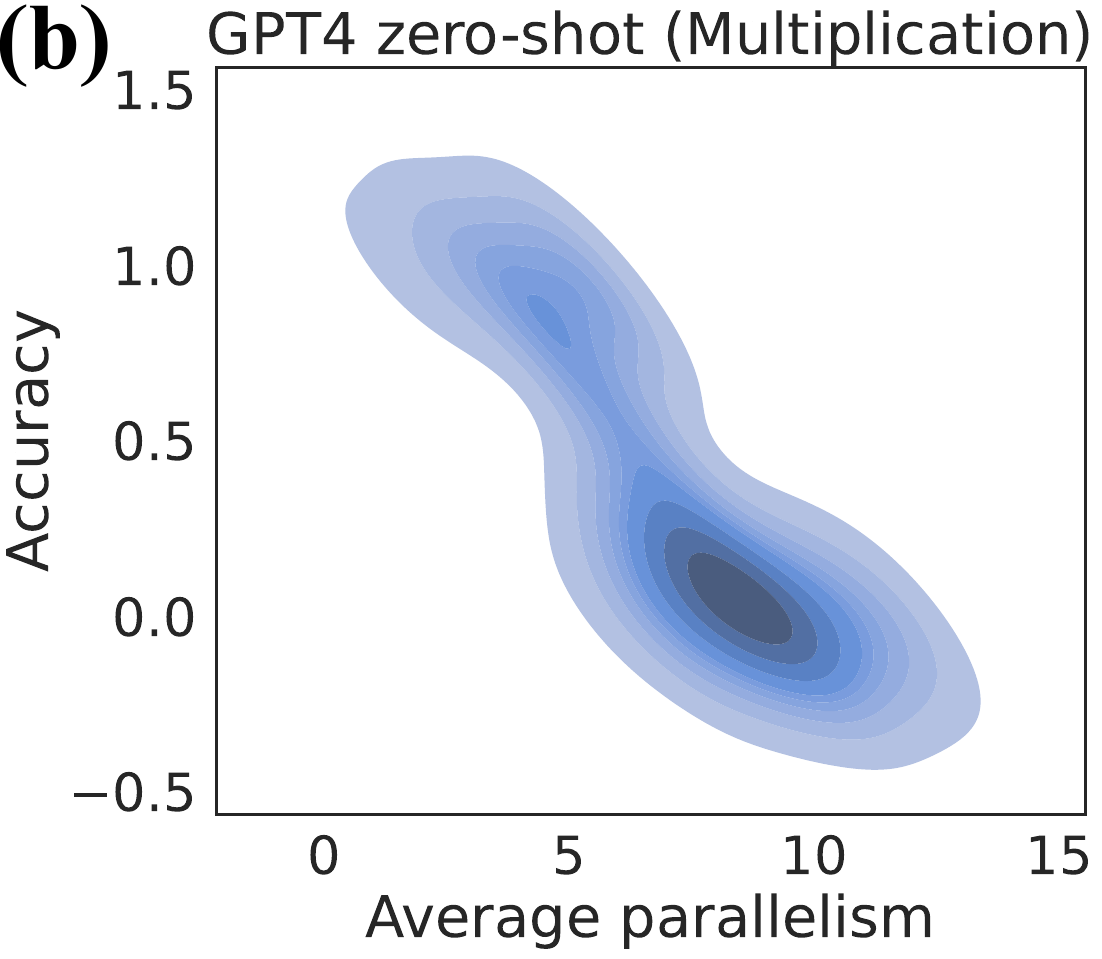}
   \end{minipage}
    \caption{Compositionality evaluation on \acp{llm}. (a) \textbf{Zero-shot accuracy}. The axes indicate the sizes of problems (number of digits in multiplication, quantity of houses and traits in puzzle-solving, and the length of sequences in dynamic programming tasks). As the complexity of a task escalates, measured by the problem size, the accuracy of Transformers dwindles nearly to zero. (b) \textbf{Average parallelism} negatively correlates with accuracy. Figure adopted from \citet{dziri2023faith}.}
    \label{fig:compositionality}
\end{figure}

The study by Dziri (2023) \citep{dziri2023faith} investigates the issue of compositionality in Transformer-based models from a graph perspective, with a particular focus on \acp{llm}. It reveals that these models struggle with three key compositional tasks: multi-digit multiplication, logic grid puzzles, and classic dynamic programming. The empirical evidence suggests that Transformers resolve these tasks not by developing systematic problem-solving skills, but rather by transforming complex, multi-step compositional reasoning into more manageable, linearized subgraph matching. \cref{fig:compositionality} depicts the zero-shot accuracy of models on the three representative tasks. While the models may show satisfactory results on relatively simpler instances of the problem, accuracy drops to near zero as task complexity increases.

\subsection{Other Empirical Results}
In this section, we delve into additional empirical evidence that challenges the notion of \acp{llm} being a form of artificial general intelligence.

\subsubsection{Shortcut Learning}
\begin{figure}[!ht]
    \centering
    \includegraphics[width=1.0\textwidth]{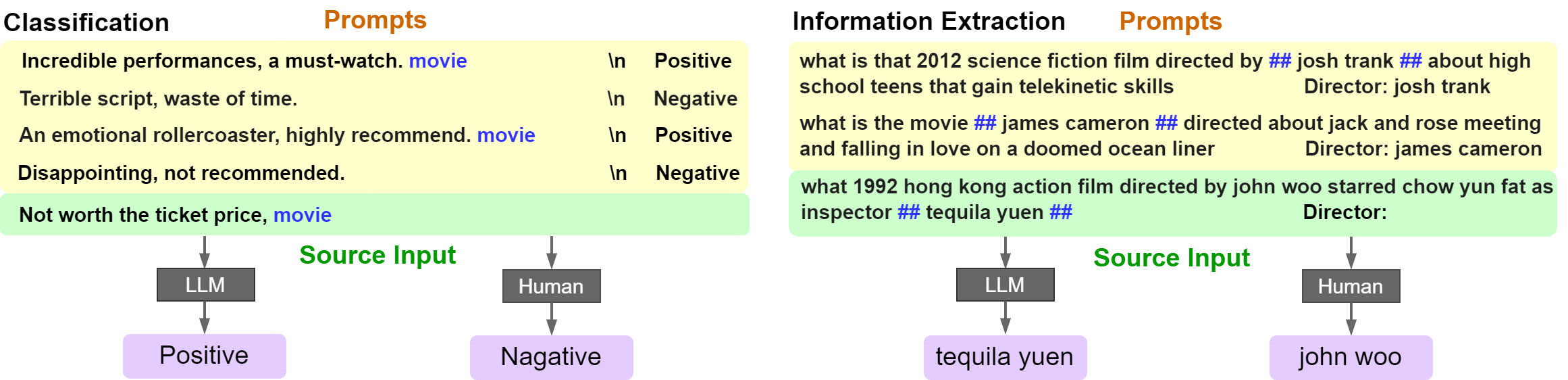}
    \caption{The text illustrates two instances of shortcut learning within the context of in-context learning. The figure on the left presents shortcuts identified during a sentiment classification task, where ``movie'' acts as a trigger word. Conversely, the figure on the right puts forth shortcuts noticed in an information extraction task, using ``\#\#'' as the trigger sign. As per the depiction in these figures, it can be noted that \acp{llm} tend to latch onto these embedded shortcuts to draw conclusions, which unfortunately results in incorrect predictions. In stark contrast, human participants seem to disregard such shortcuts. Figure adopted from \citet{tang2023largelazy}.}
    \label{fig:shortcut}
\end{figure}

\begin{figure}[!ht]
    \centering
    \includegraphics[width=0.4\columnwidth]{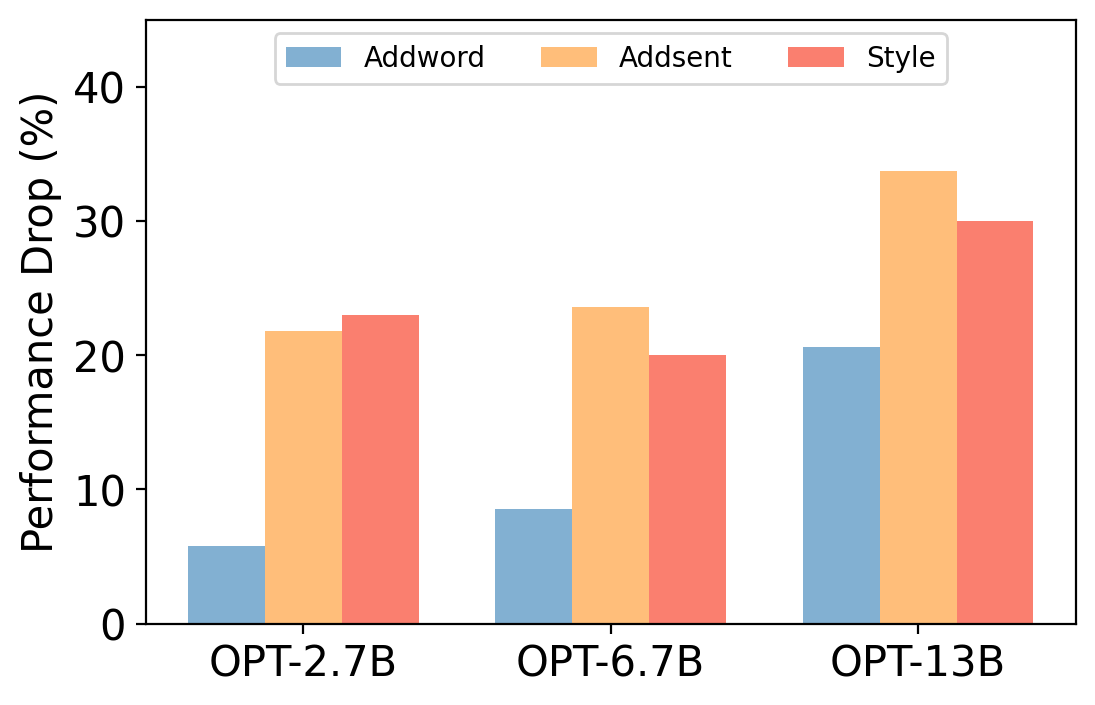}
    \caption{ Three types of shortcut triggers: words, sentences, and text styles. A noticeable decline in performance on the dataset SST2 has been observed in three \acp{llm}: OPT-2.7B, OPT-6.7B, and OPT-13B. The study identifies that these \acp{llm} tend to depend on shortcuts for downstream tasks, resulting in a significant performance drop when tested against an anti-shortcut dataset. Interestingly, the research has also uncovered the inverse scaling phenomenon --- larger models experience a more pronounced performance dip compared to their smaller counterparts. Figure adopted from \citet{tang2023largelazy}.}
    \label{fig:shortcut-perf}
\end{figure}

As a statistical learning model, \acp{llm} are also subject to the deficiency of shortcut learning. \citet{tang2023largelazy} discover that \acp{llm} rely on shortcuts or spurious correlations during in-context learning, behaving as lazy learners. Even worse, large models that empirically show better performance on a range of tasks are more likely to exploit these shortcuts. \cref{fig:shortcut} shows two examples of shortcut learning in \acp{llm}. In both of the examples, the \acp{llm} are adversely affected by the trigger words, and become trigger word detectors, rather than a classifier or an extractor. Systematic experimental evaluation in \cref{fig:shortcut-perf} further demonstrates that larger models experience a bigger performance drop.

\subsubsection{Creativity}
\begin{figure}[!ht]
\centering
\includegraphics[width=\textwidth]{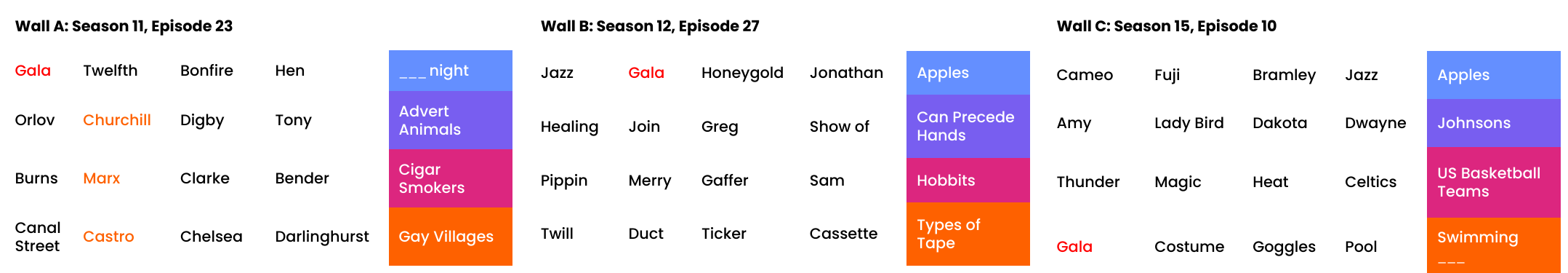}
\caption{Examples of \textit{Only Connect} walls with ground-truth groupings (rows) and connections (last column). 
\textit{Red herrings} include orthographically same words (e.g., \textcolor{red}{\textit{Gala}}) in different connected groups (\textit{\underline{Gala} night, Apples, Swimming \underline{gala}}) across walls. In Wall A (left), words \textcolor{orange}{\textit{Churchill, Marx, Castro}} provide misleading stimuli inducing plausible fixation on historical figures within the wall. Figure adopted from \citet{naeini2023large}.}
\label{fig:creativity}
\end{figure}

\citet{yiu2023imitation} approach \acp{llm} evaluation from a novel perspective: they argue that modern \acp{llm} are more like efficient imitation engines, copying existing knowledge from large corpora of data, but lack the capacity to design new tools and discover novel causal structures --- tasks that young children can easily do. Parallelly, another research conducted by \citet{naeini2023large} seeks to measure creativity in \acp{llm} quantitatively where researchers introduced a new dataset based on the game \textit{Only Connect}. The task for the model is to correctly categorize various words into four groups having hidden semantic coherence. The complexity of the task is increased by introducing distractors, known as \textit{red herrings}, which serve as misleading cues. See \cref{fig:creativity} for examples of the task. \cref{tab:creativity-gpt} showcases how GPT models fare on these creative tasks compared to human performance. The results reveal that \acp{llm} fall significantly short of humans in their performance, illustrating a stark difference.

\begin{table}[!ht]
\centering
\resizebox{0.8\textwidth}{!}{%
\begin{tabular}{@{}llcccccc@{}}
\toprule
                  &   \# In-context Examples      & WD     & FMS    & ARI    & AMI    & \# Solved Walls & \# Correct Groups \\ \midrule
GPT-3.5-turbo     & 0-shot  & $93.5$ & $16.7$ & \ \ $8.8$  & $10.4$ & $0$             & \ \ $47$              \\
                  & 1-shot  & $85.0$ & $32.5$ & $17.2$ & $20.1$ & $1$             & $105$             \\
                  & 3-shot  & $81.6$ & $35.9$ & $20.2$ & $23.4$ & $1$             & $143$             \\
                  & 5-shot  & $81.9$ & $36.4$ & $20.8$ & $24.1$ & $1$             & $141$             \\
                  & 10-shot & $82.5$ & $35.5$ & $19.8$ & $22.9$ & $1$             & $132$             \\
GPT-4             & 0-shot  & $73.9$ & $43.2$ & $\mathbf{29.1}$ & $\mathbf{32.9}$ & $5$             & $258$             \\
                  & 1-shot  & $\mathbf{73.0}$ & $\mathbf{43.3}$ & $\mathbf{29.1}$ & $32.7$ & $\mathbf{7}$             & $\mathbf{268}$             \\
                  & 3-shot  & $74.7$ & $42.7$ & $28.4$ & $32.0$ & $6$             & $246$             \\
                  & 5-shot  & $73.9$ & $42.8$ & $28.5$ & $32.2$ & $5$             & $248$             \\
                  & 10-shot  & $74.9$ & $41.6$ & $27.0$ & $30.6$ & $4$             & $238$             \\ \midrule
Human Performance &         & --     & --     & --     & --     & $285$ / $494$          & $1405$ / $1976$           \\ \bottomrule
\end{tabular}%
}
\caption{Results on the Only Connect tasks using \acp{llm}. WD: Wasserstein Distance. FMS: Fowlkes Mallows Score. ARI: Adjusted Rand Index. NMI: Normalized Mutual Information. \textbf{Bold}: best scores. Table adopted from \citet{naeini2023large}.}
\label{tab:creativity-gpt}
\end{table}

\subsubsection{Inverse Scaling}
Inverse scaling is an unexpected phenomenon that goes against our conventional understanding of artificial intelligence. Essentially, it reveals that for some tasks, bigger models may perform worse. In a recent study \citep{mckenzie2023inverse}, this inverse scaling effect was observed across 11 different datasets. The analysis from this study suggests that \acp{llm} tend to fall into several common pitfalls. Firstly, they have a bias towards repeating sequences they have previously memorized rather than following new instructions. Secondly, they frequently mimic undesirable patterns present in their training data. Thirdly, they are easily misled by deceptive information, often taking the easier route rather than accurately processing complex tasks. Lastly, they can be easily influenced by misleading demonstrations. Therefore, while it's tempting to think that larger models would inherently yield better performance, this isn't always the case. Further research is needed to understand and overcome these issues.

\subsection{Issues with the Current Evaluation Methodology}
In this section, we discuss the potential issues with the current evaluation methods. The evaluation metrics may significantly impact our perception of the capability of the \acp{llm}. \citet{schaeffer2023emergent} present an alternative explanation for the so-called emergent abilities: it is the choice of metric, rather than the fundamental changes in model behaviors, that makes us feel \acp{llm} suddenly become powerful. Simply put, for a non-linear metric, say $x^n$, sparsely sampled points on the curve make it appear that emergent behaviors happen; however, for a linear metric, such observation will be missing. Another issue is that the massive internet-scale training datasets for \acp{llm} may potentially cover the datasets used for later evaluation given that these evaluation sets are generally sourced from the internet and highly accessible. As now training sources are unavailable, the notion of generalization becomes even more vague and it becomes impossible to tell if a model really learns an underlying function, or simply retrieves it from its memory. Such non-transparency hinders genuine and reliable evaluation.

\section{Our View on Artificial General Intelligence}
It is no doubt that \acp{llm} can serve as a helpful assistant for humans --- a more personalized encyclopedia that is user-friendly through natural language conversation. However, we argue that there is still a huge gap between \acp{llm} and \ac{agi}. To lay the groundwork for subsequent discussions, we first need to clarify our understanding of \ac{agi}. There are diverging and even contradictory views on \ac{agi} which makes it difficult to find a generally accepted definition \citep{chollet2019measure,goertzel2014artificial}. In this case, we adopt a descriptive approach rather than a prescriptive one; that is, we try to extract several characteristics of \acs{agi} and present them in a coherent line of argument instead of giving a rule-based definition that presupposes correctness. There are four traits that we ascribe to \acs{agi} including

\begin{itemize}
    \item Agents can perform \textit{infinite tasks}. 
    \item Agents are \textit{autonomous to generate new tasks} in a given context. 
    \item Agents are propelled by a \textit{value system}, serving as a fundamental architecture for the generation of tasks.
    \item Agents possess a \textit{world model}, which represents the real world and guides their interaction with the world.  
\end{itemize}

To investigate \ac{agi} from a behaviorism viewpoint, we propose that generally intelligent agents should be able to perform infinite tasks in the dynamic physical and social space. Otherwise, if we set a threshold of the number of tasks that indicates the realization of \acs{agi}, it will always be questionable how this threshold is selected. If an agent is not generally intelligent when it completes $N$ tasks, there is no reason to believe that it will magically possess general intelligence once it completes $N+1$ tasks. A long checklist of specific challenging tasks is useful in terms of assessment of agent performance, like how teachers use students' scores on tests to evaluate their learning performance, but completion of specific tasks alone will not be equal to possessing general intelligence just like students' scores can not be used to stand for their true learning ability. By referring to infinite tasks, our intention is not that an agent should be omnipotent like Superman to be capable of anything. In addition, we believe that generally intelligent agents should be able to generate previously undefined new tasks in the specific context, which is similar to students learning how to learn. 

Autonomous agents capable of generating new tasks require two underlying mechanisms. Firstly, agents need a driving force that motivates the task-generation process. For example, Darwin's theory of evolution defines two instincts for all species to act upon the world, that is, to survive and reproduce \citep{popper1978natural}, which are encoded in our genes that preserve the human species. The later evolution of humankind has witnessed a diversity of values, \eg, altruism, honesty, beauty, courage. Each individual is driven by a complex value system that is shaped by his ongoing interaction with the physical and social world. A similar idea of a value system can be incorporated to create generally intelligent agents and served as an engine for agents to generate appropriate new tasks based on a predefined value system. In this case, artificial intelligence can be aligned via value alignment instead of a predefined step-by-step instruction for tasks. Secondly, agents need a world model that entails grounded representations of the real-world and implicit physical laws such as the causal chain and social norms \citep{ha2018world}. It is like a LEGO play. While the world model contains different types of bricks (the object representations) plus how they can be connected via each other (the physical laws), the value system selects an ideal, \eg, a castle, among all the other possibilities for agents to build, and the process of turning bricks to a LEGO castle requires agents to continually generate new tasks, \eg, pick up which brick to connect to an existing node, based on the current building progress. 

\section{The Unity of Knowing and Acting}\label{sec:unity}

To achieve the \ac{agi} with the grouding ability and the properties that we desire, we argue that mere knowing or acting alone is insufficient to cultivate genuinely intelligent agents.

Following Wang Yang-Ming's philosophy of knowledge as action \citep{wang1963instructions}, we argue that knowing and acting shall be deeply integrated in a learning system, where the intelligent agent actively performs actions as a way to both form comprehensive representation for the real-world objects it is interacting with, such as tactic feedback, representation from different views, and even sound, and, more crucially, to explore the surrounding environment, crystallize the knowledge from trials and errors, and to a greater extend, generalize the knowledge from acting to novel situations.

In the following, we discuss our view on the multi-faceted integration of knowing and acting from the two perspectives, \ie,
\begin{itemize}
    \item Active interaction with real-world objects provide more comprehensive signals for establishing concept representation.
    \item Knowledge is not well-learned with only passive input but shall be supplemented with experience; even unknown in the beginning, actions from repetitive trials and errors could lead to new knowledge.
\end{itemize}

\subsection{Interaction for Concept Learning}
Imagine learning a new concept as simple as a ``cup'' as a baby. Not everything ``shaped as a small and round container'' is called a cup. They may also be referred to as a ``bowl'' or a ``dust bin''. Cup is neither only shaped as a small and round container and ``\textit{with a handle}''. Some teacups do not have handles. More importantly, however, cups are usually served as a tool for \textit{drinking} liquids. One may also consider using cups for \textit{transporting} or \textit{holding} liquids or other items that fit the space.

As noted in the example, actions play a pivotal role in understanding a concept. We therefore argue that the fundamentals to fully grasp a specific concept shall include not only appearance or geometry features, but, more critically, involve the functionality and affordance of a concept one can interact with as well. Such a learning process shall be inherently multimodal: during the interaction process to learn the concept of cup, we not only see what a cup looks like when we play around with it, but we also sense the temperature of the liquids it contains, the weight the cup exerts on our hand, and the feeling of quenching our thirst when drinking. While the sensory impulses are hard to fully capture using current sensors, we believe that the widely adopted learning paradigm of learning without any interaction and with only static bi-modal input of vision and language is far from enough for one to understand a new concept.

Existing \acp{llm} behave as if a large database of established concepts with language-hashing ability and may even tell you how to use a cup for potted plants. However, for relatively new concepts, we note that they still fall short compared to humans and perform no more than statistically correlating symbols in textual corpora \citep{jiang2023mewl}, lacking in understanding multiple aspects of the concept. We argue that the absence of such interactive behaviors and accompanying sensory input consists of part of the missing pieces towards ideal general intelligence, without which the agent has no way of associating perceptual observation with effects from actions, let alone the functionality and the affordance of new concepts. 

The general problem for interactive concept learning could be formulated in a reinforcement learning framework. However, compared to existing reinforcement learning problems, concept learning shall not be task-specific or goal-driven, unlike achieving high scores in Atari \citep{mnih2015human}, navigating in an environment \citep{savva2019habitat}, or completing a language instruction \citep{shridhar2020alfred}. In some sense, concept learning should be more ``unsupervised'', as contrasive learning does for representation learning \citep{chen2020simple}. We expect the goal instantiated in interactive concept learning to be more closely related to children's inherent desire to explore or similar to curiosity-driven objectives.

To facilitate agents with human-level concept understanding, we envision a realistic meta-verse (AGI-verse, or Tong-verse), supporting far-richer interactive behaviors with objects than existing platforms. Agents in the meta-verse not only receive visual observation and textual explanation, but also can take actions with objects in the environment, or play with them, recursively apply existing concepts to new concepts, and potentially discover novel usage of a concept that is rarely encountered in the real world. Ideally, such actions also create sound effects and provide haptic feedback. Off-the-shelf assets for similar interactive environments, such as Habitat \citep{savva2019habitat} and Behavior \citep{li2023behavior}, are still poised for learning for specific tasks, with insufficient interactive action space and realistic effects. 

Going slightly beyond concept learning, we believe the foundation for the success of interactive concept learning should also serve to facilitate tool using \citep{zhu2015understanding,zhu2020dark}. With proper composition of existing concepts and their properties, we also hope that the learning mechanism would give rise to tool creation, a hallmark of human-level intelligence.

\subsection{Crystallizing Knowledge from Action}

\citet{gopnik2000detecting} propose the task of \textit{Blicket} detection that well captures the essence of turning experience of trials and errors into knowledge and how it helps generalization.

The series of experiments were initially designed for probing children's causal learning mechanism and were found to be strikingly similar to modern scientific discovery. \citet{gopnik2000detecting} introduced a special device called a Blicket machine to children subjects. The Blicket machine has a unique mechanism: if a Blicket is put on top of it, the machine will be activated, flashing and making sound. During the experimentation phase, the subjects were shown a series of experiments for compositions of objects, demonstrating the Blicketness of some of them. And then, the children were allowed time for \textit{exploratory play} with the objects. They could freely compose the objects and put the composition on top of the Blicket machine to better understand Blicketness of all objects. After the process, the subjects would be asked questions, such as which object was a Blicket and given a composition of objects that activated or did not activate the machine, how to inactivate or activate it.

While it is noted that older children generally perform better in this task, we also realize from human experiments that despite that Blicketness in the beginning is uncertain for all objects, better-performing subjects make informed trials for quick disambiguation. The interactive trial-and-error process significantly improves final problem-solving; with only passive observation and no active interaction, the uncertainty will only remain. Furthermore, for questions regarding intervention, \eg, what would happen if an object were added or removed, subjects with intensive interaction with the machine show notable superiority.

\citet{xu2022est} develop a computational interactive environment for the Blicket detection problem based on the static setup from \citet{zhang2021acre}. Their newly introduced \textit{EST} environment mimics the classic Blicket detection experiment but intentionally simplifies sensorimotor control by abstracting it out into a discrete space of object selection. Their experimental results from computational methods show that existing learning methods, including completely \ac{llm}-based ones, fare pathetically worse than naive and inefficient heuristics search.

While there has been success using iterative prompting, supplying \acp{llm} with both actions and effects and iteratively running the process, for interactive learning \citep{wang2023voyager,zhu2023ghost}, we note that the model is still largely based on internet-scale knowledge colossus on a specific subject \citep{fan2022minedojo}. Yet, with the presumably low exposure of Blicket on the internet, the \acp{llm} become no better than random actors. The results suggest that existing methods rely largely on existing passively provided data and are simply unable to crystallize new knowledge from novel phenomena by active interaction.

We believe that artificial general intelligence shall possess the ability to quickly act to resolve ambiguity and turn the experience from successful and unsuccessful interventions into their knowledge on how to interact with the environment, instead of being only fed with data passively, with no capacity to demystify novel situations through interaction and knowledge acquisition.

The reinforcement learning setup inherently supports acting and learning. However, existing reinforcement learning problems for common sense intelligence are more perceptual than reasoning, which requires instant response rather than complex System-2 computation on-the-fly. We hope that a challenging reasoning problem based on interaction with the environment will emerge, serving as a testbed for evaluating how the community performs on turning experience of trials and errors into knowledge and further use the knowledge to perform additional everyday tasks. Besides, it is also unresolved how to abstract from existing knowledge and apply them as general principles to novel situations. Knowledge abstraction, knowledge accumulation, and knowledge application should be the critical processes in realizing such systems. We believe that a realistic meta-verse mentioned above shall also serve as an important factor in building the living environment for an agent to play and learn.

\section{Discussion and Future Directions}

In this work, we review existing failure cases for \acfp{llm} and refute the reckless claim that \acp{llm} represent ``sparks of artificial general intelligence'' \citep{bubeck2023sparks}. Analysis from both careful benchmarking and empirical observation suggests that \acp{llm} may be a good database that hashes language queries, but far from the general intelligence demonstrated in humans. Besides, deficiency in evaluation also casts doubt on the validity of the results on existing web-sourced datasets, as the largest of the \acp{llm} may have already been trained on them.

We further present our view on artificial general intelligence and propose the unity of knowing and acting, a factor critical for living agents yet paradoxically missing in the acclaimed intelligent \acp{llm}. In our view, the unity of knowing and acting could serve at least to help concept learning and knowledge acquisition. 

Following the discussion, we point out three future directions for advances in artificial general intelligence research.

\subsection{Transparent Evaluation}
As the dataset size becomes increasingly bigger, the critical issue of generalization is gradually ignored; so long as the model performs ``well'' on the test set, it is considered good. However, the good performance may stem from training on the testing data, and in fact the model does not really understand the problem. Close-sourced models like GPT \citep{openai2023gpt} further cloud interpretation for evaluation. As evaluation datasets are usually sourced from the internet and the \acp{llm} are trained on internet data, we argue that a new way of evaluation is in desperate need that can make sure limited data leakage from the internet to warrant true generalization.

\subsection{Affordance-rich Interactive Environments}
As mentioned in \cref{sec:unity}, a founding component for knowing and acting is a meta-verse. The meta-verse should ideally support rich affordance that allows an agent to play with objects for concept learning, providing feedback of multi-modality, including more than just vision and language. The meta-verse should also support a variety of reasoning tasks, covering tasks regarding knowledge acquisition and abstraction, such as instruction following, planning, abduction, and induction. Of particular interest should be tasks without extensive internet data exposure, in order to solidify the argument that the agent learns from interaction with the environment rather than retrieving from given knowledge. 

\subsection{Unifying Knowing and Acting}
We argue that a cognitive architecture shall be developed to integrate knowing and acting. Despite of success from reinforcement learning in a narrow domain of tasks, a general mechanism for knowledge as action should transcend pure data-driven approaches for generalization in knowledge abstraction, knowledge accumulation, and knowledge application. Besides, it still remains an open problem on how to formulate existing knowledge and incorporate off-the-shelf knowledge into new knowledge discovery. We hope that a cognitive mechanism can be scalable enough and seamlessly combine knowledge-driven and data-driven benefits.

In the end, while we acknowledge the great practical advances \acp{llm} have brought to the community, we strongly believe that they do not represent artificial \textit{general} intelligence and hope that this article serves as inspiration for the research community towards the ultimate goal.

\bibliographystyle{apalike}
\bibliography{references}

\end{document}